\documentclass[sigconf,authorversion]{acmart}

\usepackage{array}
\usepackage{pifont}
\usepackage{xcolor}
\usepackage{listings}
\usepackage{dirtree}
\usepackage{tikz-cd}
\usepackage{xspace}
\usetikzlibrary{positioning, arrows.meta}
\usepackage[noabbrev,capitalize]{cleveref}  
\usepackage{autonum} 
\usepackage{enumitem}
\usepackage{caption}
\usepackage[skip=0.5ex, belowskip=1ex,
                labelformat=brace,
                singlelinecheck=off]{subcaption}

\newcommand\YAMLcolonstyle{\color{red}\mdseries}
\newcommand\YAMLkeystyle{\color{black}\bfseries}
\newcommand\YAMLvaluestyle{\color{blue}\mdseries}

\makeatletter

\newcommand\language@yaml{yaml}

\expandafter\expandafter\expandafter\lstdefinelanguage
\expandafter{\language@yaml}
{
  keywords={true,false,null,y,n},
  keywordstyle=\color{darkgray}\bfseries,
  basicstyle=\YAMLkeystyle,                                 %
  sensitive=false,
  comment=[l]{\#},
  morecomment=[s]{/*}{*/},
  commentstyle=\color{purple}\ttfamily,
  stringstyle=\YAMLvaluestyle\ttfamily,
  moredelim=[l][\color{orange}]{\&},
  moredelim=[l][\color{magenta}]{*},
  moredelim=**[il][\YAMLcolonstyle{:}\YAMLvaluestyle]{:},   %
  morestring=[b]',
  morestring=[b]",
  literate =    {---}{{\ProcessThreeDashes}}3
                {>}{{\textcolor{red}\textgreater}}1     
                {|}{{\textcolor{red}\textbar}}1 
                {\ -\ }{{\mdseries\ -\ }}3,
}

\lst@AddToHook{EveryLine}{\ifx\lst@language\language@yaml\YAMLkeystyle\fi}
\makeatother

\newcommand\ProcessThreeDashes{\llap{\color{cyan}\mdseries-{-}-}}

\newcolumntype{C}{>{\centering\arraybackslash}m{2.5cm}}
\newcommand{\rev}[1]{#1}

\newcommand{\ie}{\textit{i.e., }}

\lstdefinestyle{mystyle}{
    backgroundcolor=\color{gray!5},
    basicstyle=\ttfamily\small,
    breakatwhitespace=false,
    breaklines=true,
    captionpos=b,
    commentstyle=\color{green!40!black},
    keywordstyle=\color{blue},
    numberstyle=\tiny\color{gray},
    stringstyle=\color{orange},
    numbers=left,
    numbersep=5pt,
    showspaces=false,
    showstringspaces=false,
    showtabs=false,
    tabsize=2
}

\AtBeginDocument{%
  \providecommand\BibTeX{{%
    \normalfont B\kern-0.5em{\scshape i\kern-0.25em b}\kern-0.8em\TeX}}}

\copyrightyear{2024}
\acmYear{2024}
\setcopyright{licensedothergov}\acmConference[ACM REP '24]{ACM Conference on Reproducibility and Replicability}{June 18--20, 2024}{Rennes, France}
\acmBooktitle{ACM Conference on Reproducibility and Replicability (ACM REP '24), June 18--20, 2024, Rennes, France}
\acmDOI{10.1145/3641525.3663648}
\acmISBN{979-8-4007-0530-4/24/06}

\begin{document}

\title[\rev{MLXP: A Framework for Conducting Replicable Experiments in Python}]{\rev{MLXP: A Framework for Conducting\\ Replicable Experiments in Python}}

\author{Michael Arbel}
\affiliation{%
  \institution{Univ.\ Grenoble Alpes, Inria, CNRS, Grenoble INP, LJK}
  \city{ 38000 Grenoble}
  \country{France}
}

\author{Alexandre Zouaoui}
\affiliation{%
  \institution{Univ.\ Grenoble Alpes, Inria, CNRS, Grenoble INP, LJK}
  \city{ 38000 Grenoble}
  \country{France}
}

\renewcommand{\shortauthors}{Arbel and Zouaoui}

\begin{abstract}
Replicability in machine learning (ML) research is increasingly concerning due to the utilization of complex non-deterministic algorithms and the dependence on numerous hyper-parameter choices, such as model architecture and training datasets.
Ensuring reproducible and replicable results is crucial for advancing the field, yet often requires significant technical effort to conduct systematic and well-organized experiments that yield robust conclusions.
Several tools have been developed to facilitate experiment management and enhance reproducibility; however, they often introduce complexity that hinders adoption within the research community, despite being well-handled in industrial settings. 
To address the challenge of low adoption, we propose MLXP, an open-source, simple, and lightweight experiment management tool based on Python, available at  \href{https://github.com/inria-thoth/mlxp}{https://github.com/inria-thoth/mlxp}. MLXP streamlines the experimental process with minimal practitioner overhead while ensuring a high level of reproducibility.
\end{abstract}



\begin{teaserfigure}
  \centering\includegraphics[width=.8\textwidth,page=3]{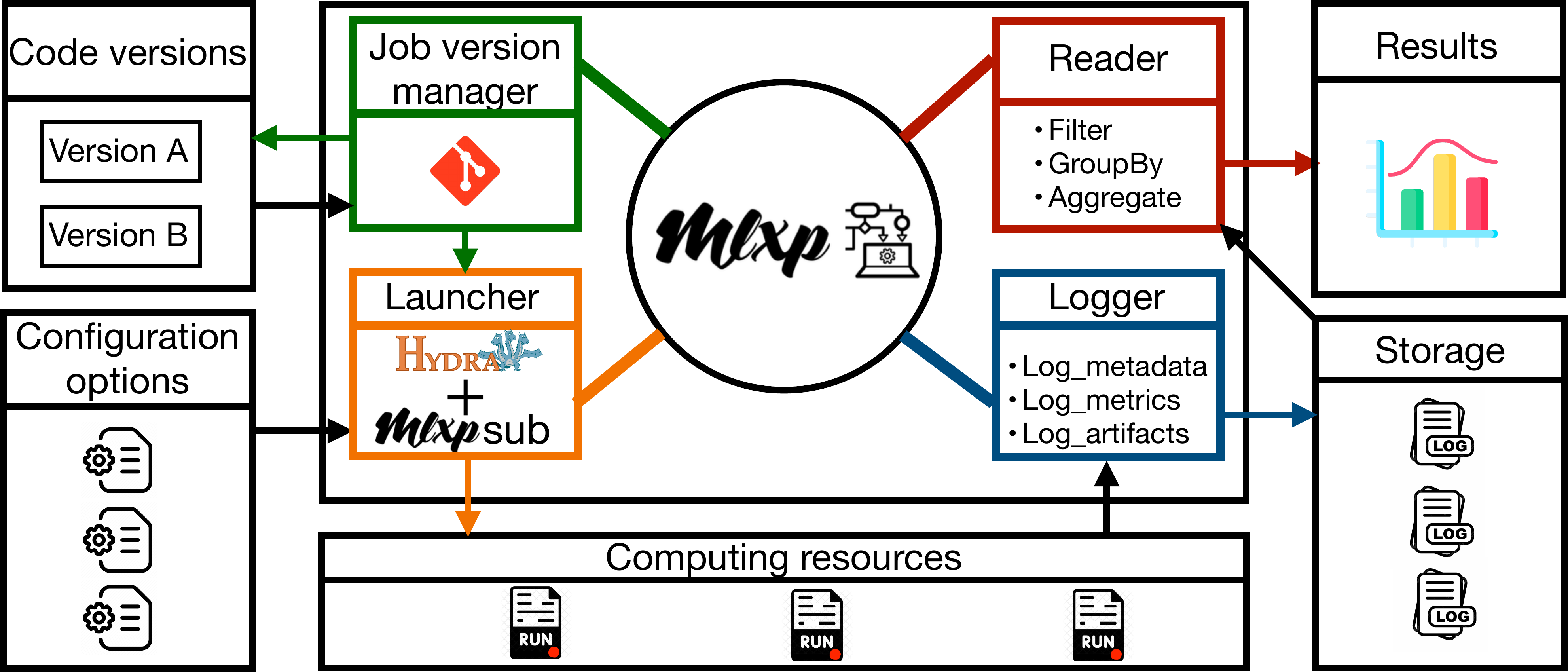}
  \caption{Overview of MLXP and its components}
  \label{fig:teaser}
\end{teaserfigure}

\maketitle

\section{Introduction}
In recent years, the field of Machine Learning (ML) has experienced rapid advancements in both research and practical applications, ranging from improved algorithms to novel use cases in various industries.
Recent breakthroughs, such as Large Language Models, have been projected to yield a potential annual market value of trillions of dollars, according to a report by McKinsey \citep{mckinseyreport}, underlining their extensive applicability and  the significant economic opportunities they present.

Many machine learning models, such as Large Language Models for text and speech generation \citep{team2023gemini} and autonomous driving systems \citep{GrigorescuJFR20}, are extremely challenging to train as they rely on sophisticated algorithms and demand extensive computational resources and access to vast databases. 
As such, these models substantially differ from traditional software systems governed by deterministic rules. ML models are constructed using stochastic methods, introducing randomness through factors such as the choice of initial weights or the order of data presentation to the model. This randomness makes it challenging to reliably assess the performance of a given method from a single execution of the algorithm. 
Additionally, training often involves complex optimization procedures with numerous hyperparameters like scheduling, momentum, and weight decay, necessitating a systematic search for optimal configurations. 
These hyperparameters, coupled with architectural choices and data selection, compound the complexity, demanding meticulous organization and execution of potentially numerous experiments to ensure reproducibility and replicability which are essential for making progress in research.

A significant portion of data scientists and ML practitioners tend to rely on customary solutions, often characterized as ad hoc and lacking a robust methodology for experiment management \citep{hill2016trials,vartak2016modeldb}. 
As a result, these customary choices may introduce friction in the experimentation process, potentially undermining reproducibility and replicability \citep{HutsonScience18}. 
For instance, the ability to systematically test ideas and conduct extensive grid searches over hyperparameters is crucial for impartially comparing different methods. Yet, this capability can be compromised when using customary solutions, which often result in cumbersome code. Consequently, the resulting complexity imposes a psychological burden and an increased workload on practitioners when conducting multiple experiments, thereby introducing frictions to reproducibility and replicability.

Careful handling of experiments, coupled with a robust methodology and efficient tools for experiment management, is necessary to ensure the reproducibility and replicability of results \citep{Idowu:2022}. 
Recent years have seen a rise in experiment management tools, such as MLflow, Wandb, Neptune.ai, Comet.ml, and Sacred, which  provide extensive functionalities aimed at ensuring the reproducibility and traceability of ML workflows. These tools primarily focus on systematic tracking and logging of various ML assets (datasets, models, hyperparameters, etc.), versioning them, and visualizing and comparing the results of several experiments. 
Leveraging these tools can significantly enhance reproducibility and traceability. 
However, this advantage comes with a considerably steep learning curve and the need to (re)organize ML project code-bases to effectively utilize them, potentially limiting their adoption within the broader scientific community \citep{mora2021traceability,ormenisan2020implicit}. 
Specifically, these tools often introduce their own set of APIs, utilities, and concepts that researchers must familiarize themselves with and learn to use. 
Consequently, the learning and integration burden associated with these tools can be prohibitive in research contexts, where projects prioritize agility and flexibility over deployment considerations that enforce strict workflows and conventions well suited to industrial settings. Simple experiment management tools with a low entry barrier, tailored to meet the scientific demands of flexibility and frictionless reproducibility, are lacking in the current landscape. 
There is a need for tools that can offer such a frictionless experience starting from configuration, launching, logging to versioning and exploitation of the results while requiring minimal intrusiveness in practitioners code-base to increase adoption by the research community.

We introduce MLXP (\cref{fig:teaser}), an open-source Python framework designed to streamline experiment management, ensuring minimal friction during experimentation and result exploitation while upholding essential reproducibility standards with minimal intrusion. 
Python is widely recognized as the language of choice for machine learning, given its extensive array of machine-learning-related packages like Jax \citep{jax2018github} or Pytorch \citep{paszke_pytorch_2019}, and its widespread adoption in the research ecosystem.  
Therefore, we have chosen Python as the native language for MLXP, enabling seamless integration with the code-bases of the majority of machine learning projects. 
MLXP builds upon Hydra \citep{yadan_hydra_2019}, a popular ML framework that has garnered increasing attention from the ML research community thanks to its intuitive and concise syntax for configuring and locally launching multiple experiments. 
MLXP extends Hydra's capabilities by 
\rev{ simplifying }
  multiple job submission to job schedulers, \rev{and allowing} logging, version management, searching, and post-processing experiments, \rev{ to ensure } transparent experiment management. 
The result is a user-friendly system that prioritizes simplicity and empowers users to conduct rigorous experiments with confidence, fostering a culture of replicability in data science research. %

\paragraph{Paper Organization} \Cref{sec:related_work} discusses existing tools for experiment management, highlighting their strengths and limitations in the scientific research context. \Cref{sec:challenges} addresses challenges encountered when conducting ML experiments for research and proposes a set of requirements that an experiment management tool needs to satisfy to mitigate these challenges. \Cref{sec:Hydra} provides background information on Hydra, while \Cref{sec:MLXP} introduces MLXP and its core components. In \cref{sec:applications}, we provide examples of research projects that leverage MLXP to easily perform replicable experiments. Finally, we conclude this paper with \cref{sec:conclusion}.

\section{Related work}\label{sec:related_work}

\subsection{Best practices for replicability in ML} 
Rising concerns over replicability of experiments in ML has spurred the development of new best practices aimed at ensuring a high level of reproducibility and replicability in research findings. 
Recent initiatives include various documentation frameworks for ML models \citep{GebruDatasheet18,MitchellFAT19} and checklists \citep{mlreprochecklist}, which outline the essential assets (e.g., datasets, code, experimental results) necessary to reproduce ML models. 
Recent work emphasized the importance of utilizing version management tools tailored for ML code, data, and environments to ensure a high level of reproducibility in machine learning \citep{Chen:2022d,IdowuSEIP21,AmershiSEIP19,BarrakSANER21}. 
Furthermore, major AI conferences such as NeurIPS, ICML, and AAAI regularly host reproducibility workshops and encourage researchers to independently verify published research results in an effort to incentivize the adoption of best practices for reproducibility in scientific work \citep{Sinha:2023}.  
Despite these efforts, best practices for reproducibility and replicability remain less prevalent in the research community compared to other organizational settings, as highlighted in a study by \citet{Serban:2020}, which surveyed over 300 practitioners from various organizations. The study revealed low adoption rates for crucial practices like automated/systematic hyper-parameter optimization and model selection, which are vital for replicating ML research findings and ensuring their validity.
We attribute the limited adoption of such practices in research institutions to a mismatch between the complexity required to enforce them and the need for flexibility and simplicity in research environments. 
Our proposed framework aims to address this mismatch by providing a lightweight tool that integrates key best practices, assisting practitioners in conducting replicable experiments.

\subsection{ML experiment management tools}
An emerging category of tools aims to tackle the complexities associated with managing ML experiments,  
\rev{such as MLFlow \citep{chen_developments_2020}, WandB \citep{biewald_wandb_2020}, NeptuneML, Deep-water \citep{ferenc2020deep}, Runway \citep{tsay2018runway}, and ModelKB \citep{gharibi2019automated, gharibi2019modelkb}.
}
Most of these tools primarily focus on furnishing functionalities for storing, tracking, and versioning assets across different experiment runs to address concerns like reproducibility 
\citep{isdahl2019out,tatman2018practical} and traceability \citep{mora2021traceability}. 

With the emergence of experiment management tools, conducting systematic comparisons between them has become imperative to identify to what extend they can ensure reproducibility. 
Several studies aimed to compare these tools based on criteria deemed critical for reproducibility \citep{IdowuSEIP21,Idowu:2022,isdahl2019out,weissgerber2019mapping},  
such as the granularity of tracked/logged information and the level of intrusiveness required to access this information. 
Even though they offer very extensive functionalities such as tracking versioning different types of assets, studies such as \citet{Idowu:2022}, concluded that these management tools still lack maturity compared to their traditional software engineering counterparts which may limit their adoption by a wider community. 
Factors contributing to their lack of maturity include friction and overhead incurred during usage due to the necessity of code instrumentation for tracking assets \citep{mora2021traceability,ormenisan2020implicit}. 
Specifically, many of the most popular experiment management tools like MLFlow,  typically lack solutions for handling interactions with computing resources, such as easily launching multiple jobs to a job scheduler or managing delayed job executions, both crucial for frictionless replicability. 
Even when these solutions are provided, (ex. Wandb) they often require mastery of the proposed submission tools and introduce an overhead for practitioners. 
Additionally, while these frameworks enable the exploitation of results from multiple experiments, users may witness performance degradation or scalability issues when dealing with a large number of experiments or extensive datasets without careful adjustments.
In this work, we introduce a simple and lightweight experiment management tool that addresses some of these limiting factors by significantly simplifying the experimental process with minimal user overhead while ensuring a fundamental level of reproducibility.

\section{Design goals and principles} \label{sec:challenges}

\rev{Verifying replicability of data scientific conclusions  requires both ensuring  reproducibility of experiments and testing the effect of hyper-parameter choices and randomness on such conclusions. 
This requirement often implies repeating experiments multiple times under different conditions. 
Without suitable tools, 
such a task can quickly become challenging as it may require producing a large amount of additional code for configuring, submitting experiments and storing and post-processing results. This added code may also lack the necessary flexibility to adapt to evolving experimental requirements (e.g. additional hyper-parameters due to a modification of a method under development). 
The frictions created by this coding overhead, can constitute an obstacle to replicability. 
Starting from this observation, we advocate for the following principle when designing an experiment manager:}

\textit{An experiment manager tool can facilitate replicability of a data scientific experiment when it enables reusing the same code that produces a single result for producing potentially thousands more variant results.}

\rev{From the above principle, we identify four design goals that we propose to pursue when developing an experiment manager that facilitates replicability:}

\begin{enumerate}[label=\textbf{(G.\arabic*)},ref=\textbf{G.\arabic*}]
    \item \label{chall:config} \rev{Simple configuration and submission of multiple jobs.}
    \item \label{chall:boiler} \rev{Automated and flexible logging of information.} 
    \item \label{chall:hpc} 
    \rev{Automated code and job version management.}
    \item \label{chall:results} \rev{Integrated analysis tools of the results.}
\end{enumerate}

\paragraph{Goal \ref{chall:config}.} 
\rev{Without specific tools tailored for configuring and submitting experiments, practitioners often need to produce additional code simply for repeating the same experiment using different choices of hyper-parameters. This additional code can take the form of nested for-loops in command-line languages such as} \verb+bash+.  \rev{This approach is particularly cumbersome when several hyper-parameters need to be tested and can thus constitute a barrier to systematic testing of the robustness of experimental results. An ideal candidate experiment manager that facilitates replicability of data scientific experiments should allow configuration and submission of multiple jobs to a HPC job scheduler without requiring additional coding effort.}

\paragraph{Goal \ref{chall:boiler}.}
\rev{Storing relevant information about a given experiments in a systematic manner is crucial for ensuring the replicability and reproducibility of these experiments. 
This entails recording metadata, including hyper-parameters, environment details and relevant experimental outputs in a manner that allows: 
1- recovering the inputs provided to the code for a given experiment, and 
2- reproducing the same recorded outputs given the recovered inputs.   
Manually recording such information is error-prone, as it necessitates additional code development to ensure that each experiment is stored in a separate location and that the record contains sufficient information to guarantee reproducibility.
An experiment manager should simplify such a process by automating logging of relevant information while providing enough flexibility to practitioners for deciding what additional information needs to be stored.}

\paragraph{Goal \ref{chall:hpc}.} 
\rev{Tracking changes made to the code-base that led to significant performance improvement can be challenging, especially when multiple experiments are submitted continuously, while the code is still under development.
In addition, the use of HPC systems can result in additional challenges related to delayed and asynchronous job execution. 
Jobs submitted to an HPC might be launched after several, possibly defective, changes have been introduced into the code-base if the computing resources were busy at the time of job submission. This can lead to erroneous job executions, based on a unreliable code. 
Automatically handling code and job versioning to avoid these pitfalls is valuable in an experiment manager to ensure traceability and reliability of the experimental results and thus contributing to the replicability of experiments.}

\paragraph{Goal \ref{chall:results}.} 
\rev{In the context of replicability, large amounts of experiments might be needed to verify the robustness of scientific conclusions. Consequently, practitioners are faced with numerous experimental results that demand rigorous analysis. 
Such analysis often requires additional coding effort for extracting results, appropriately formatting them and post-processing them. Reducing such effort can be accomplished using an experiment manager equipped with integrated analysis tools that can  streamline operations including results extraction, filtering, grouping, and aggregation. }

\rev{Our proposed solution, MLXP aims at achieving the above design goals by providing four core functionalities described in \cref{sec:MLXP}: Launching, Reading, Job versioning and Reading. Next, we describe Hydra's framework on which MLXP relies on for concisely configuring experiments. }

\section{Background on Hydra}\label{sec:Hydra}
Hydra \citep{yadan_hydra_2019} \rev{is a popular open-source framework for configuring experiments hierarchically.
Its main features are: 
1- the ability to extract configurations (e.g. hyper-parameters) for an experiment from a default configuration file,  
2- the ability to override these extracted configurations from the command line interface (CLI) without modifying the default configuration file. 
These features result in drastic simplification to the code since passing complex options becomes modular.}

{\bf Using Hydra for configuring experiments.} \rev{\cref{fig:hydra_multi_listing} shows how Hydra can be used to extract default configuration options from a configuration file and \emph{sequentially} execute multiple runs with different choices for these options that are overridden from the command line.  \cref{code:hydra_config} provides an example of a} \verb+YAML+ \rev{configuration file} \verb+config.yaml+ \rev{located in a sub-directory} \verb+configs+ \rev{of the current working directory and containing default parameter values. Hydra can then extract these values and provide them as a dictionary-like object} \verb+DictConfig+ \rev{to the main function of a} \verb+python+ \rev{file} \verb+main.py+ \rev{(\cref{code:hydra_main}). 
This is achieved simply by adding a decorator} \verb+@hydra.main+ \rev{to the main function and executing the command} \verb+python main.py+ \rev{from the CLI.} \rev{The power of Hydra lies in the ability to concisely \emph{override} the default values from the CLI to execute multiple jobs. \cref{code:hydra_multirun} shows how to \emph{sequentially} execute four jobs, each  corresponding to a choice of values for the tuple of parameters} (\verb+seed+,\verb+lr+) \rev{amongst the cross-product set $\{1,0\}\times \{1.0,0.1\}$. Hence, a potentially large number of experiments can be sequentially submitted from a single command.}

{\bf Configuring experiments without Hydra.} 
\rev{\cref{code:argparse} illustrates an alternative approach, without Hydra, for running the same experiments using} \verb+argparse+ \rev{package. \cref{code:argparse_main} shows the modifications needed to the python file} \verb+main.py+ \rev{for defining all optional parameters. }
\rev{These modifications to the code introduces a coupling between configuration parameters and the python code. Thus, any change to the configuration requires explicit modification  to the code. Moreover, running multiple instances of the script with different options is a cumbersome operation as underlined by the bash script in \cref{code:argparse_batch} which uses nested loops.}

\begin{figure}[htbp]
\begin{subfigure}{0.45\textwidth}
\begin{lstlisting}[language=yaml, frame=single, style=mystyle]
seed: 0
lr: 10.
trainer:
 num_epoch: 10
model:
 num_units: 100
\end{lstlisting}
\caption{Example of a 'config.yaml' configuration file}
\label{code:hydra_config}
\end{subfigure}

\begin{subfigure}{0.45\textwidth}
\begin{lstlisting}[language=Python, frame=single, style=mystyle]
import hydra

@hydra.main(config_path='./configs',
            config_name='config.yaml')
def task_function(DictConfig: cfg)->None:
    print(cfg)

if __name__ == "__main__":
  task_function()
\end{lstlisting}
\caption{Example of a 'main.py' Python script using Hydra}
\label{code:hydra_main}
\end{subfigure}

\begin{subfigure}{0.45\textwidth}
\begin{lstlisting}[language=bash, frame=single, style=mystyle]
$ python main.py seed=0,1 lr=1.0,0.1 --multirun
\end{lstlisting}
\caption{Running multiple jobs using Hydra}
\label{code:hydra_multirun}
\end{subfigure}

\caption{Configuring experiments using Hydra.}
\label{fig:hydra_multi_listing}%
\end{figure}

\begin{figure}[htbp]
\begin{subfigure}{0.45\textwidth}
\begin{lstlisting}[language=Python, frame=single, style=mystyle]
import argparse

def task_function(args):
    print(args)

if __name__ == "__main__":
  parser = argparse.ArgumentParser()
  parser.add_argument("-s","--seed",type=int,default=0)
  parser.add_argument("-l","--lr",type=float,default=10.0)
  parser.add_argument("-e","--epoch",type=int,default=10)
  parser.add_argument("-u","--units",type=int,default=100)

  args = parser.parse_args()
  task_function(args)
\end{lstlisting}
\caption{Example of a 'main.py' Python script using argparse}
\label{code:argparse_main}
\end{subfigure}
\begin{subfigure}{0.45\textwidth}
\begin{lstlisting}[language=bash, frame=single, style=mystyle]
for seed in 0 1
do
    for lr in 1.0 0.1
    do
        python main.py --seed $seed --lr $lr
    done
done
\end{lstlisting}
\caption{Bash script to use multiruns with argparse}
\label{code:argparse_batch}
\end{subfigure}
\caption{Configuring experiments without Hydra.}
\label{code:argparse}
\end{figure}

\rev{The simplifications to the code induced by Hydra and illustrated in \cref{code:argparse,fig:hydra_multi_listing} has made it a popular tool in the ML community with several successful research projects relying on it \citep{suvorov_resolution-robust_2022,takamoto_pdebench_2022,sodhani_multi-task_2021}. }

{\bf Going beyond Hydra.} 
\rev{While Hydra is an advanced tool for configuring experiments, it is not designed to be a complete experiment management tools as it does not provide support for advanced logging, job versioning and result exploitation. 
Nevertheless, our proposed solution builds on Hydra's capabilities to provide a complete experiment management tool for ML practitioners. Since the additional functionalities of MLXP are orthogonal to Hydra's core capabilities, it is natural to depart from its original code-base. } 

\section{MLXP}\label{sec:MLXP}

MLXP (Machine Learning eXperimentalist for Python) is an open-source Python package whose source code and documentation are available in \href{https://github.com/inria-thoth/mlxp}{https://github.com/inria-thoth/mlxp} and \href{https://inria-thoth.github.io/mlxp/}{https://inria-thoth.github.io/mlxp/} and that can be easily installed using PyPi  (see \href{https://pypi.org/project/MLXP/}{https://pypi.org/project/MLXP/}). 
MLXP builds upon the powerful Hydra framework to offer a \rev{ lightweight tool} for easily managing multiple experiments in Python to ensure their reproducibility and replicability. 
As an open-source package, MLXP facilitates  experiment launching, logging, and \rev{efficient} result exploitation. Key components include automated job launching using Hydra and hierarchical configuration files, logging of experiment outputs along with metadata, 
automated code and job version management, seamless multi-job submission to a HPC job scheduler, and intuitive result exploitation capabilities including querying results, grouping and aggregation operations. 
Using these components we address the challenges described in \cref{sec:challenges}. The end result is an easy-to-use system that prioritize transparency and empowers users to conduct rigorous experiments with confidence, fostering a culture of replicability and reproducibility in data science research. We provide an overview of MLXP in \cref{sec:Overview} to illustrate its simplicity, before discussing its major components. 

\subsection{Overview and usage}\label{sec:Overview}

MLXP provides a drop-in decorator \verb+mlxp.launch+ that extends  Hydra's  decorator \verb+hydra.main+  and allows to dynamically pass a context object \verb+ctx+ to a main function \verb+task_function+ defined inside a python script. 
This decorator is designed to seamlessly integrate with existing code-bases, requiring minimal adjustments to the user's code while offering maximum compatibility and ease of integration (ex: see the \verb+main.py+ script in \cref{code:launch}). 
\begin{lstlisting}[language=Python, frame=single, style=mystyle, caption={Example of a Python script ('main.py') using MLXP}, label={code:launch}]
import mlxp

@mlxp.launch(config_path='./configs')
def task_function(ctx: mlxp.Context)->None:
  config = ctx.config
  logger = ctx.logger

  print("Hello World!")

if __name__ == "__main__":
  task_function()
\end{lstlisting}
The context object \verb+ctx+ provides two essential elements for launching the experiment: (1) a hierarchical experiment configuration object \verb+ctx.config+ similar to Hydra's \verb+DictConfig+ structure, that provides information from a default configuration files and can be overridden in the command line, and (2) a \verb+logger+ object to store the outputs/results of the experiment in a directory automatically created by MLXP and that is unique to a particular run. 

\begin{figure}
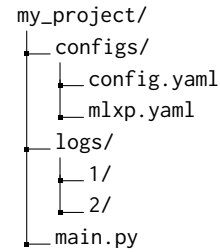

    \centering
 \begin{minipage}{0.2\textwidth}
    \dirtree{%
.1 my\_project/.
.2 configs/.
.3 config.yaml.
.3 mlxp.yaml.
.2 logs/.
.3 1/.
.3 2/.
.2 main.py.
}
\end{minipage}
    \caption{Project directory structure}
    \label{fig:directory_structure}
\end{figure}

Using MLXP requires only defining two reserved directories for the project: a configuration directory and a log directory. 
For instance, \cref{fig:directory_structure} shows an example of a project directory structure which contains these two directories in addition to the main python script \verb+main.py+:. 
The configuration directory \verb+configs+ contains two \verb+yaml+  files \verb+config.yaml+ and \verb+mlxp.yaml+ from which MLXP extracts the default configuration for the project. 
The \verb+logs/+ directory, whose location can be customized by the user, is created by default by MLXP under working directory and contains separate sub-directories, each storing information provided to the \verb+logger+ object during a specific run. 
These sub-directories are named after their order of execution (1, 2, ....) so as to ensure a unique identifier.  
Finally, executing Python scripts using MLXP follows the same procedures as Hydra, meaning that regular Python command-lines as simple as \verb+python main.py+ can be used. 

\subsection{Launching}\label{sec:launching}
Launching experiments \rev{easily} is paramount in research and development environments. Hydra provides an excellent mechanism for executing multi-runs with ease, allowing users to \rev{simply} pass multiple options through the command line. MLXP takes this a step further by seamlessly integrating with job schedulers using \verb+mlxpsub+ command. This integration enables users to not only orchestrate and configure experiments \rev{easily} but also to submit them directly to a job scheduler without manual intervention. The resulting automation  enhances productivity and ensures consistent and reliable execution of experiments across different computing environments.

\subsubsection{The mlxpsub command}
MLXP provides a shell command (\verb+mlxpsub+) which leverages the multi-run functionalities of Hydra for \rev{simple} submission of multiple runs using a single bash script. It is compatible with most of the available job schedulers, such as \verb+SLURM+, \verb+TORQUE+, \verb+SGE+, \verb+OAR+, \verb+MWM+ and \verb+LSF+, and simply requires to specify the jobs' options in the main script using the syntax defined by the scheduler. 
\cref{code:script} illustrates the simplicity by which several jobs can be specified transparently in a few lines of code using a single bash script. There, it is assumed the user has access to a \verb+SLURM+ job scheduler.  Therefore, the job specifications, such as the duration of each job, the number of tasks and the number of cpus per task, follows the scheduler's syntax as indicated by lines starting with the string \verb+#SLURM+. 
\rev{ Finally, following Hydra's syntax, the last line creates 4 different jobs, each corresponding to an execution of the python script} \verb+main.py+ \rev{using a particular value for the tuple of parameters} (\verb+seed+,\verb+lr+) \rev{amongst the cross product set $\{1,0\}\times \{1.0,0.1\}$. }
\vfill 
\begin{lstlisting}[language=bash, frame=single, style=mystyle, caption={Example of a Bash script ('script.sh') using MLXP}, label={code:script}]
#!/bin/bash

#SLURM --time=1-00:10:00
#SLURM --ntasks=1
#SLURM --cpus-per-task=1

python main.py  lr=10.,1. seed=1,2
\end{lstlisting}
Simply launching the script in \cref{code:script} using \verb+mlxpsub script.sh+ command will automatically create 4 different jobs, each corresponding to a tuple of options provided to the python script \verb+main.py+, which are then submitted to a scheduler's queue. 
By contrast, launching the script in \cref{code:script} directly using a scheduler's submission command (here \verb+sbatch script.sh+) will create a single job which successively executes 4 different runs with their corresponding tuple of options. 

\subsubsection{Alternatives}
Without the \verb+mlxpsub+ command, one would have to create a \textit{meta} script that generates a script similar to \verb+script.sh+ for each option choices and submits it using \verb+sbatch+ command. 
Such approach does not exploit the multi-run functionality provided by Hydra and introduces clutter in the job submission code. 
Other popular alternatives, such as \verb+submitit+ package, allow a simplified job submission to a SLURM scheduler, but requires adapting the Python code for using it and does not support command-line submission natively. 
Note, however, that a plugin version of \verb+submitit+ for Hydra allows command-line submission,  although it is specific to SLURM and  requires passing the scheduler's options in a format that is different from the scheduler's native one.
On the other hand, \verb+mlxpsub+ is versatile as it is compatible with several schedulers and transparent because it simply relies on the scheduler's native syntax, with which the user is supposedly already familiar.

\subsection{Logging}\label{sec:loggin}
Keeping track of information relative to an experiment/run is crucial for replicability. 
The MLXP's logger component provides logging capabilities that (1) automatically handles the creation of log directories unique to each run, (2) systematically stores \verb+metadata+ relative to a run, such as the configurations used during execution, information about the resources used,  code version, etc and (3) simply allows the user to log additional information about the run such as various metrics, checkpoints or artifacts. 
All log directories created using MLXP have a similar structure as the one shown in \cref{fig:log_directory_structure}. Below, we discuss key requirements that MLXP satisfies to allow maximal replicability and ease of use.  

\subsubsection{Automatic creation of unique log directories} 
When running multiple experiments in parallel, storing the result of each one separately guarantees that no interference can occur. MLXP ensures the each new experiment has a unique location that is resolved using a simple internal mechanism for keeping track of previous runs in a given log directory. Such a mechanism 
increments the total number of existing log directories by one to assign a \verb+log_id+ to each new run.  The new run directory is then named after the new \verb+log_id+. When submitting multiple jobs to a job scheduler using \verb+mlxpsub+ command discussed in \cref{sec:launching}, these jobs are executed asynchronously. A naive approach would be prone to a large I/O concurrency when resolving the \verb+log_id+ for each job. 
MLXP avoids these issues altogether by sequentially assigning a \verb+log_id+ to each job and creating its corresponding log directory before submitting these jobs to a scheduler. Then upon execution, each job is forced to store its outputs in its assigned log directory. As a result, MLXP allows to easily submit a large number of jobs without worrying about clashes between different jobs.

\subsubsection{\rev{Logging metadata, metrics and artifacts.} } 
\begin{figure}
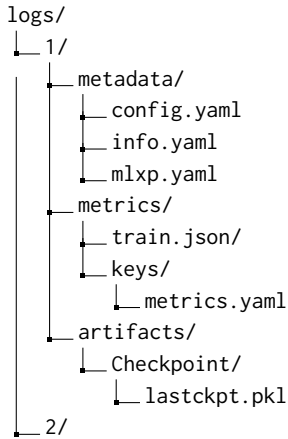

    \centering
 \begin{minipage}{0.25\textwidth}
    \dirtree{%
.1 logs/.
.2 1/.
.3 metadata/.
.4 config.yaml.
.4 info.yaml.
.4 mlxp.yaml.
.3 metrics/.
.4 train.json/.
.4 keys/.
.5 metrics.yaml.
.3 artifacts/.
.4 Checkpoint/.
.5 lastckpt.pkl.
.2 2/.
}
\end{minipage}
    \caption{Log directory structure}
    \label{fig:log_directory_structure}
\end{figure}

Storing information such as all the configuration options used to run experiments are basic requirements for replicability that MLXP handles automatically. There, the object \verb+ctx.config+ is stored in a file \verb+config.yaml+ under the \verb+metadata+ directory of a given run (see \cref{fig:log_directory_structure}). 
Moreover, information about the execution of the run, such as its 'status' (ex: COMPLETED, FAILED, RUNNING) or MLXP's configuration options are stored in separate files (\verb+info.yaml+ and \verb+mlxp.yaml+) to allow a precise  monitoring of experiments. For instance, a user might want to handle results coming from incomplete experiments differently to draw correct conclusions. These additional information are especially useful when several experiments are executed and need to be filtered according to their status as discussed in \cref{sec:reading}. 
These \verb+metadata+ information complement the \verb+metrics+ and \verb+artifacts+ that are stored by the user using \verb+logger.log_metrics+ and \verb+logger.log_artifacts+. The former method allows storing a list of dictionaries of scalar values in a simple \verb+JSON+ format whereas the latter one allows storing more structured objects such as images, model parameters, etc. 
Both modalities can be accessed easily thanks to the universal formats used and the simple file structure of the log directories \cref{fig:log_directory_structure}. 
Finally, due to re-allocation of computing resources, some experiment needs to be momentarily stopped and then automatically relaunched, as is the case when submitting a job to a \verb+besteffort+ queue of a scheduler. 
The ability to store generic checkpoints (in a \verb+pickled+ format) allows to safely resume the experiment in the exact state in which it was paused. This could be achieved, for instance, by calling the method \verb+logger.log_checkpoint(ckpt)+ where \verb+ckpt+ refers to any user-defined object that contains all the environment and state variables used by a given experiment. 

\subsubsection{Alternatives}
The logger provided by MLXP induces a moderate level of intrusiveness in the code as it requires the user to use call  methods such as \verb+log_metrics+ inside the code-base. 
This approach is quite common to most experiment management tools and does not constitute a strong constraint in the context of scientific/academic projects.  
Some tools such as ModelKB, MLFlow, and Wandb support automatic asset collection in non-intrusive ways. However, these approaches require explicit support for machine learning frameworks, such as Pytorch \citep{paszke_pytorch_2019} and SciKit Learn \citep{pedregosa2011scikit}, and appear to be constraining in some research contexts where new insights are obtained from tracking non-standard quantities of interest.

\subsection{Job versioning}\label{sec:versioning}
In typical machine learning research projects, there is a constant iteration between code development and experimental validation. This iterative process provides flexibility to test new ideas and refine them based on partial empirical results. While this agile development approach can accelerate scientific research, it also increases the likelihood of erroneous conclusion about a given method due to misreporting experimental results. 
Misreporting is likely to occur when data from different code versions are combined potentially altering the interpretation of the experiments. \rev{Although} unintended, this constitutes a breach of the scientific method and can result in significant wasted time and resources. 
MLXP proposes a systematic way to version the experiments and to guarantee they were obtained using a specific version of the code-base. Before describing MLXP's versioning mechanism, we first illustrate how a lack of experiment versioning can have unintended consequences when using HPC clusters and job schedulers. 
\subsubsection{Parallel job submissions without versioning.}
\begin{figure}
\centering
\includegraphics[width=0.45\textwidth,page=1]{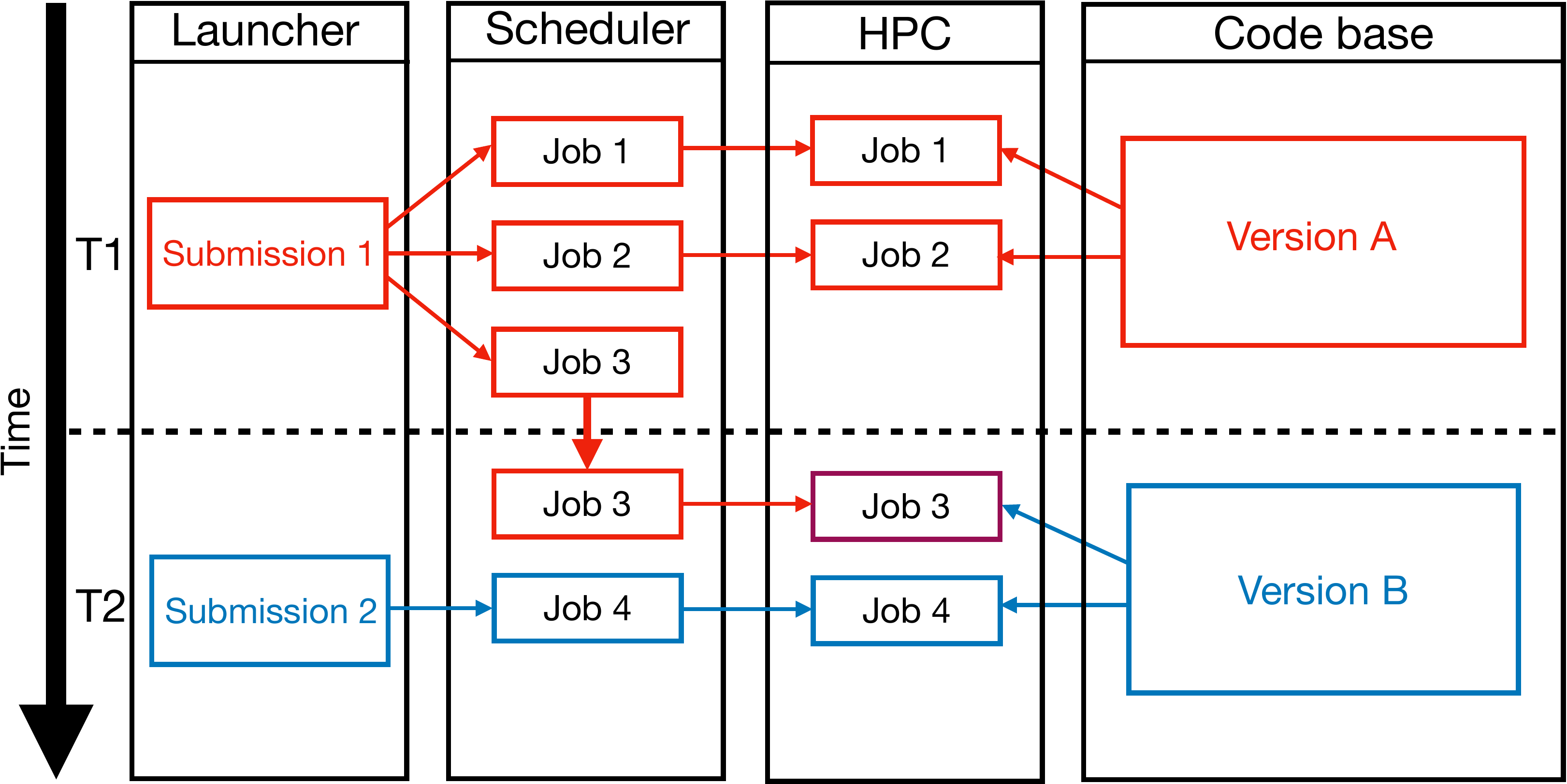}
\caption{Execution of non-versioned jobs}
\label{fig:codebase_jobs}
\end{figure}

A typical situation the users can be confronted to in practice arises when submitting multiple jobs to a busy HPC cluster. 
In such cases, jobs are placed in a scheduler's queue until the required resources become available. 
Consequently, jobs in the queue may run asynchronously, with an uncertain start time, while still relying on a shared code-base. 
This code-base is likely to undergo changes even before the jobs are executed, as illustrated in \cref{fig:codebase_jobs}. 
There, three jobs are first submitted to a scheduler at time $T1$ and intended to use the current code version (version $A$). However, due to resource constraints, only the first two jobs are executed with the correct version at $T1$. 
The third job (job 3) is deferred to time $T2$, at which point a new code version (version $B$) becomes available, and a new job (job 4), intended to use the latest code version, is submitted to the queue. 
Consequently, both jobs 3 and 4 are executed using the current code version (version $B$). Had sufficient resources been available at $T1$, job 3 would have used version $A$ of the code. Thus, depending on resource availability, the same job submitted to an HPC cluster may be executed using different code versions.  Next, we discuss how MLXP addresses this replicability issue. 
 
\subsubsection{MLXP's version manager}
MLXP introduces a version manager that builds on \verb+Git+ to address the replicability issue that might arise due to a delayed execution of jobs. 
The version manager operates by versioning the code with \verb+Git+ and associating all jobs submitted to a scheduler at a specific time with a \rev{version} of the code based on the latest commit available. 
When the requested resources become available for a job, it is executed from \rev{a copy of the code corresponding the code version associated to the job} rather than the current code-base, which may have changed in the meantime. 
To prevent redundant copies of the code, \rev{at most a single copy is created for each} \verb+Git+ \rev{commit and is uniquely named after the } commit hash. 
Thus, \rev{as long as the code does not change,} only a single \rev{copy} of the code is created, even if multiple jobs are submitted.

\begin{figure}
\centering
\includegraphics[width=0.45\textwidth,page=2]{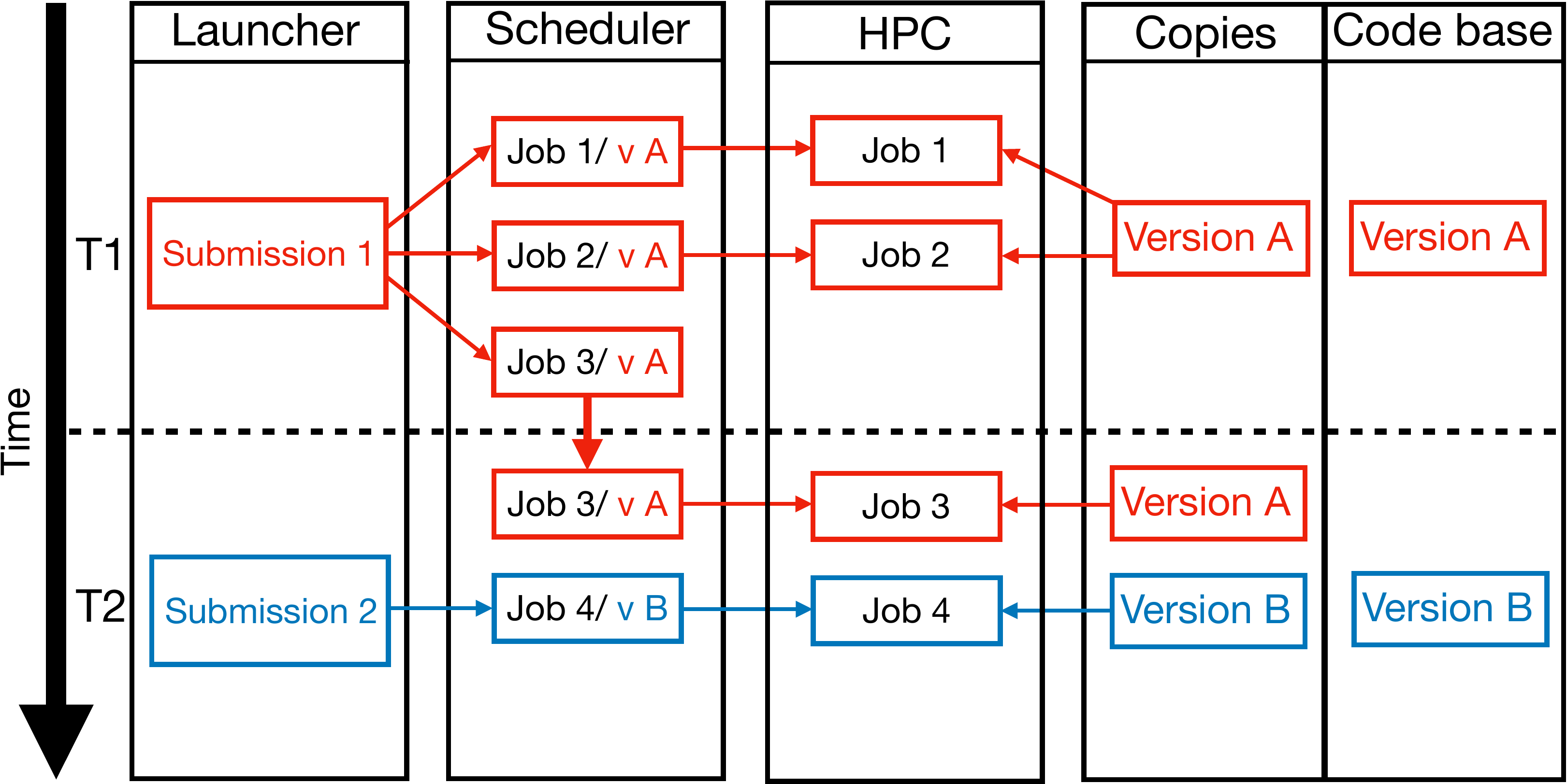}
\caption{Execution of versioned jobs}
\label{fig:codebase_jobs_version}
\end{figure}

\cref{fig:codebase_jobs_version} illustrates how linking jobs to code \rev{versions} ensures that each job is executed using the intended code version, irrespective of changes in the current version of the main code-base.
In this scenario, even though job 3 was initially linked to version A, it will still be executed using that same version from the backup \rev{version}, despite its execution being delayed until time $T2$, when the current code-base has evolved to version B.
This enhanced traceability of experiments increases their replicability in the context of large HPC clusters. Furthermore, as a byproduct of this traceability, it becomes easier to pinpoint potential bugs that could have been introduced at subsequent versions of the code. This is the case, if for instance, in \cref{fig:codebase_jobs_version}, the new job (job 4) is expected to give the same results as (job 3) even though it is using a new version of the code (version B). Any discrepancy in the results, can alert the user about a potential bug that could either be introduced in version B or that was present in version A and was inadvertently fixed later. 

\subsubsection{Code version checking}
The effectiveness of the proposed strategy relies upon a rigorous versioning of the code-base using \verb+Git+, which might not be always guaranteed, especially in a context of fast-paced code development and limited exposure to version control practices. To overcome these limitations, MLXP offers optional version control checks that can be used in an interactive manner. When these are enabled and upon launching a script that uses MLXP, an interactive session is created where MLXP checks if the current \verb+Git+ repository containing the code-base has untracked files or uncommitted changes. It then asks the user if they want to add untracked files and create an automatic commit. Only after that, the jobs are created dynamically based on the user's choices and are submitted to a job scheduler when available. While this procedure is optional, it can guarantee optimal versioning and traceability of the experiments, which can save a lot of time and resources especially when each experiment is computationally demanding. 

\subsection{Reading}\label{sec:reading}
The MLXP components discussed so far focused on facilitating the submission of multiple jobs, ensuring their traceability and correct execution through rigorous versioning, and maintaining records of all pertinent information about each experiment, including configuration details, results, and run status. 
With these components, users can \rev{easily} submit a large number of experiments, resulting in a substantial amount of data. Properly designed tools are essential for efficiently managing and exploiting this data.
In an effort to provide a \rev{complete} framework for managing experiment, we propose a \verb+reader+ component that allows easy filtering of experimental results as well as simple grouping and aggregation operations. The \verb+reader+ component also offers a lazy evaluation mechanism that avoids storing in memory potentially large amount of data produced by the experiments and only loading them when necessary. Below we describe the main functionalities of the \verb+reader+ component. 

\subsubsection{Filtering results}
Filtering results of multiple experiments using MLXP is as easy as creating a \verb+reader+ object linked to the \verb+logs+ directory of all runs and providing a query string that allows to filter results based on information stored in the \verb+metadata+ directory created by MLXP logger for each run (see  \cref{code:filtering}). 
The syntax for querying is close to Python's syntax for boolean operators and allows comparison operations ('==', '!=', '<', '>', '<=', '>='), logical operations ( and ('\&'), or ('|'), not ('$\sim$') ), and membership ('in'). It also supports operation precedence using parenthesis and follows similar  precedence rules as in Python which allows to easily combine all these comparison operations to create sophisticated queries. For ease of use, the list of keys that can be used in a query are provided by the attribute \verb+reader.searchable+ of the reader. 
\begin{figure}[htbp]
\begin{subfigure}{0.45\textwidth}
        \begin{lstlisting}[language=Python, frame=single, style=mystyle]
In [1]: import mlxp
In [2]: # Creating a reader object.
        reader = mlxp.Reader('./logs/')
In [3]: # Searching using a query string
   ... query = "info.status == 'COMPLETE' & config.optimizer.lr <= 1."
   ... results = reader.filter(query_string=query)
    \end{lstlisting}
    \caption{Reading and filtering results}
    \label{code:filtering}
    \end{subfigure}
\hfill
\begin{subfigure}{0.45\textwidth}
\begin{lstlisting}[language=Python, frame=single, style=mystyle]
In [4]: # Display results
   ...: results 
Out[4]:
   config.seed config.lr ... train.iter train.loss
0       0          1.    ... LAZYDATA   LAZYDATA
0       1          1.    ... LAZYDATA   LAZYDATA
In [5]: # Direct access 
   ...: results[0]['train.iter']
Out[5]:
    [1,2,3,4,5,6,7,8,9,10]
\end{lstlisting}
    \caption{Displaying and accessing ``lazily" evaluated results}
    \label{code:loading}
    \end{subfigure}

\begin{subfigure}{0.45\textwidth}
\begin{lstlisting}[language=Python, frame=single, style=mystyle]
In [6]: # List of group keys.
       ... group_keys = ['config.lr']
In [7]: # Grouping the results 
   ...: grouped_res = results.groupBy(group_keys)
   ...: grouped_res
Out[7]:
          config.seed ...  train.iter train.loss
config.lr
    1.          0     ...  LAZYDATA   LAZYDATA
                1     ...  LAZYDATA   LAZYDATA
\end{lstlisting}
    \caption{Grouping results}
    \label{code:grouping}
    \end{subfigure}
\hfill
\begin{subfigure}{0.45\textwidth}
\begin{lstlisting}[language=Python, frame=single, style=mystyle]
In [8]: # Creating the aggregation maps 
    ... from mlxp.data_structures.contrib.aggregation_maps import AvgStd
    ... agg_maps = [AvgStd('train.loss')]
In [9]: # Aggregating the results 
   ...: agg_res = grouped_res.aggregate(agg_maps)
   ...: agg_results
Out[9]:
      config.lr  ...   train.loss_avg
0         1.     ...   [0.03, ..., 0.001]
\end{lstlisting}
\caption{Aggregating results}
    \label{code:aggregating}
    \end{subfigure}
\caption{Example of interactive IPython shells for analyzing results using MLXP's filtering, lazy access, grouping and aggregations methods.}
\label{fig:multi_listing}%
\end{figure}
\subsubsection{Loading results} 
Once filtered, results are represented as a dataframe-like object, as shown in \cref{code:loading}, where each row represents a different run in the logs directory, whereas each column represents either a \verb+metadata+ field appearing in at least one of the runs or a \verb+metric+ that was stored by calling the method \verb+log_metrics+ of MLXP's logger object. 
While \verb+metadata+ are typically scalar quantities that can readily be stored in the dataframe-like object \verb+results+, \verb+metrics+ are typically arrays of arbitrary size that could be cumbersome to load in the dataframe for each run. 
Instead, \verb+metrics+ are ``lazily evaluated", meaning that they are simply linked to the location where they are stored and are only loaded when the user explicitly tries to access them. Hence the values of these fields are marked as ``LAZYDATA" and produce an output array whenever direct access for a particular run and metric is performed. 
Such a lazy evaluation allows a lightweight handling of large amounts of information while still retaining simplicity and flexibility. Finally, it is always possible to easily convert the MLXP dataframe-like object into a Pandas dataframe \citep{reback2020pandas} for convenience. In that case, however, ``lazy evaluation" is no longer possible. Instead, all data are loaded in memory.
\subsubsection{Grouping and aggregation}
It is often useful to group/aggregate results by values taken by some fields of the dataframe. While all these operations are possible using frameworks such as Pandas, we provide some basic support for those that is compatible with the lazy evaluation functionality of MLXP reader. 
\cref{code:grouping} illustrate how to perform a grouping operation on the dataframe given a list of group keys and using the method \verb+groupBy+ which returns a dictionary of dataframes whose keys correspond to the different group values. 
The group keys can contain several fields of the dataframe in which case the groups are indexed by a tuple of values corresponding to these fields. 
It is worth noting that the \verb+metrics+ are still lazily evaluated, which allows to perform complex grouping operations without unnecessary memory overload.   
Finally, once groups are formed, it is possible to perform simple aggregation operations. \cref{code:aggregating} illustrate how to average the field \verb+train.loss+ over members of each group using the method \verb+aggregate+ and the map \verb+AvgStd+ provided by MLXP. 
The operation returns a dataframe containing the initial group keys and the results of aggregation. Here, lazy evaluation is no longer performed since computing the aggregated result requires directly accessing the metric \verb+train.loss+.

\subsubsection{Alternatives}
MLFlow \citep{chen_developments_2020} and Weights \& Biases (W\&B) \citep{biewald_wandb_2020} provide filtering capabilities that are similar to MLXP but no direct support for grouping or aggregation operations which are usually performed using frameworks such as Pandas. 
When dealing with large amount of output data, conversion of raw experimental results into Pandas format to perform these operations can result in performance degradation and scalability issues since all results are loaded in memory. 
The fully integrated support for these operations in MLXP allows users to easily exploit and compare multiple results with a minimal code overhead and memory usage. 
\rev{These functionalities already facilitate analysis/evaluation of the results to draw  replicable conclusions: e.g. by averaging results over multiple seeds and checking robustness/sensitivity to hyper-parameters). Future developments of MLXP could provide automated analysis tools that build upon these dataframe functionalities to allow the user obtaining insights on the experimental results in a visual format (e.g. graphics).}

\section{Examples}\label{sec:applications}

In this section, we provide two examples of research projects that build upon MLXP to perform replicable experiments. 

\subsection{HySUPP \citep{rasti_image_2023}}
\begin{figure}[H]
\centering 
\includegraphics[width=0.5\textwidth]{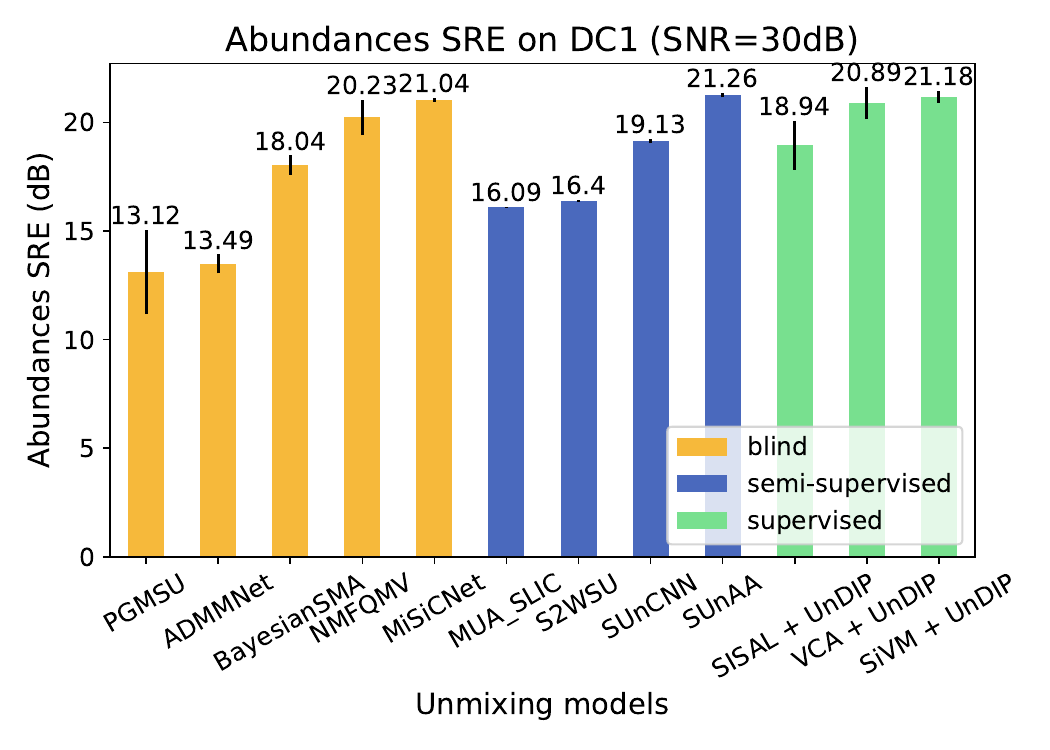}
\caption{Sub-figure originating from \citet{rasti_image_2023}}
\label{fig:hysupp}
\end{figure}

HySUPP is an open-source python toolbox for hyperspectral unmixing practitioners.
In essence, HySUPP enables seasoned researchers, curious students, and teachers to easily experiment with various hyperspectral unmixing methods with replicability in mind thanks to MLXP, as indicated in the Github repository of the code associated with the paper \href{https://github.com/BehnoodRasti/HySUPP}{https://github.com/BehnoodRasti/HySUPP}.
HySUPP leverages basic functionalities including \rev{simple} launching, logging capabilities \rev{including evaluation metrics specific to unmixing} \rev{, and} the \verb+reader+ component to easily filter experimental results before grouping and aggregating them.
Notably, the \verb+reader+ component enables end-users to create informative plots such as \cref{fig:hysupp} using minimal boiler-plate code in order to visualize the performance of several unmixing techniques, grouped by their supervision setup (\ie supervised, semi-supervised, or blind) where 5 runs have been aggregated so as to create error bars corresponding to the standard deviation from the mean value when accounting for randomness in data generation and model initialization.
An example of a command-line instruction can be found in \cref{lst:hysupp}, which highlights its intuitive usage, since users may simply change the unmixing \verb+data+, \verb+model+, or \verb+SNR+ depending on their needs.

\begin{lstlisting}[language=bash, frame=single, style=mystyle, caption={HySUPP command-line arguments}, label=lst:hysupp]
$ python unmixing.py data=DC1 model=EDAA SNR=30
\end{lstlisting}

\subsection{Benchmarking optimization algorithms}
\begin{figure}[H]
\includegraphics[width=0.45\textwidth]{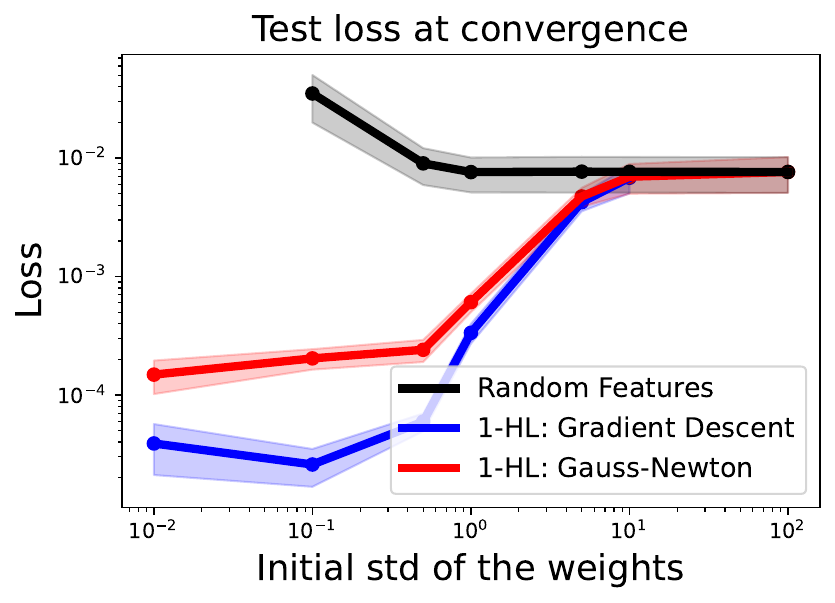}
\caption{Sub-figure originating from  \citet{arbel2023rethinking}}
\label{fig:gauss-newton}
\end{figure}

MLXP has recently been used to conduct experiments for comparing a number of optimization algorithms for deep learning. In particular, \citet{arbel2023rethinking} compared two classes of optimization methods: Gradient descent and Gauss-Newton, to provide empirical insights on the effect of these algorithms on the generalization capabilities of over-parameterized networks. There, 720 independent runs totaling in 3600 GPU hours were submitted to an HPC cluster using MLXP as indicated in the Github repository of the code associated with the paper \href{https://github.com/MichaelArbel/Implicit-Bias-Gauss-Newton}{https://github.com/MichaelArbel/Implicit-Bias-Gauss-Newton}. 
\cref{fig:gauss-newton} shows the results of a comparison between different learning algorithms depending on the standard variation of the initial parameters. Each point in the figure was obtained using $5$ independent runs each of which trains the model using a different method, totalling in $105$ independent runs.   \cref{lst:gauss-newton} shows a simple bash script that can be launched using \verb+mlxpsub+ and that submits all runs to an HPC cluster using OAR. 

\begin{lstlisting}[language=bash, frame=single, style=mystyle, caption={Bash script for job submission}, label=lst:gauss-newton]
#!/bin/bash

#OAR -l core=1, walltime=00:30:00
#OAR -t besteffort
#OAR -p gpumem>'16000'

python main.py std=0.01,0.1,0.5,1.,5,10,100\
    seed=0,1,2,3,4\
    method=RF,GD,GN\
\end{lstlisting}
Once all the runs are complete, it is possible to extract the desired results using MLXP's reader module as shown in \cref{code:gauss-newton}. This example illustrates the ease by which experiments can be systematically performed, starting from job submissions to results extraction. 
\begin{lstlisting}[language=Python, frame=single, style=mystyle, caption={Example of code for extracting results}, label={code:gauss-newton}]
import mlxp
from mlxp.data_structures.contrib.aggregation_maps import AvgStd

reader = mlxp.Reader("./logs/")
results = reader.filter(query_string="")

# Averaging over seeds 
agg_maps = [AvgStd('test.loss')]
group_keys = ['config.method', 'config.std']
agg_res = results.groupBy(group_keys).aggregate(agg_maps)
\end{lstlisting}

\section{Limitations and Future Directions}
While MLXP provides simple features for managing experiments, several future directions of improvements could result in more flexibility to the user. These improvements include, advanced analysis tools and automated reports, as well all as extended logging that supports data, model versioning and stores information about the hardware and software dependencies.

\rev{
{\bf Advanced analysis tools:} A possible direction of improvement for MLXP, is to provide advanced automated analysis tools that leverage the basic dataframe operations in MLXP to provide automated and customizable analytics. These tools could automatically measure sensitivity of some metrics to hyper-parameters, and generate automatic displays summarising them.  To facilitate this process, the MLXP dataframes can be extended to support columns representing objects such as figures produced by these analysis tools or general artifacts stored by the logger. This would require carefully handling data-types stored in the dataframe and would be the subject of future developments in MLXP.}

\rev{
{\bf Automated reports:} Currently, MLXP provides an elementary feature, the method} \verb+diff+ \rev{of the class} \verb+dataframe+, \rev{which identifies the configurations that vary amongst a list of runs. However, this feature does not directly provide insights about the outcome of individual experiments. Future developments in MLXP can combine this basic functionality with the planned advanced analysis tools to provide comparative reports for individual runs.}

\rev{
{\bf Logging hardware and software versions:} By default MLXP allows logging the versions of the python dependencies used by a run. However, results can also depend on the hardware used and versions of softwares external to python (cuda drivers, etc).}

\rev{
{\bf Versioning models and data:} Versioning any input and output of the run (such as the data and models) is important. Currently, outputs of a run can be versioned as artifacts, while inputs, such as data, can be versioned by their paths. Possible improvement could include a specific versioning method for both data and models and ensuring it to be used systematically for each experiment.}

\section{Conclusion}\label{sec:conclusion}
In conclusion, MLXP \rev{provides a simple way to handle} the complexities of managing data science experiments within a research environment. 
While MLXP currently lacks explicit support for data or model versioning, as well as interactive visualization and collaborative features, its core focus on simplicity lays a foundation for future development. Potential enhancements could include increased support for versioning and a reduction in logging intrusiveness to offer more flexibility to users. \rev{Nevertheless, current MLXP's functionalities already }
streamlines experiment launching, logging, and tracking, while facilitating efficient job submission to cluster environments and offering job versioning capabilities. With a strong emphasis on reproducibility and user-friendliness, MLXP simplifies experiment management tasks, allowing researchers to focus on their primary objectives.

\begin{acks}
This work was supported in part by the ANR BONSAI project (grant ANR-23-CE23-0012-01).	
\end{acks}

\bibliographystyle{ACM-Reference-Format}
\bibliography{MLXP}


\begin{thebibliography}{37}


\ifx \showCODEN    \undefined \def \showCODEN     #1{\unskip}     \fi
\ifx \showDOI      \undefined \def \showDOI       #1{#1}\fi
\ifx \showISBNx    \undefined \def \showISBNx     #1{\unskip}     \fi
\ifx \showISBNxiii \undefined \def \showISBNxiii  #1{\unskip}     \fi
\ifx \showISSN     \undefined \def \showISSN      #1{\unskip}     \fi
\ifx \showLCCN     \undefined \def \showLCCN      #1{\unskip}     \fi
\ifx \shownote     \undefined \def \shownote      #1{#1}          \fi
\ifx \showarticletitle \undefined \def \showarticletitle #1{#1}   \fi
\ifx \showURL      \undefined \def \showURL       {\relax}        \fi
\providecommand\bibfield[2]{#2}
\providecommand\bibinfo[2]{#2}
\providecommand\natexlab[1]{#1}
\providecommand\showeprint[2][]{arXiv:#2}

\bibitem[mck(2021)]%
        {mckinseyreport}
 \bibinfo{year}{2021 (accessed August, 2021)}\natexlab{}.
\newblock \bibinfo{booktitle}{\emph{Notes from the AI Frontier Insights from
  Hundreds of Use Cases}}.
\newblock
\urldef\tempurl%
\url{https://www.mckinsey.com/featured-insights/artificial-intelligence/
  notes-from-the-ai-frontier-applications-and-value-of-deep-learning}
\showURL{%
\tempurl}


\bibitem[mlr(2021)]%
        {mlreprochecklist}
 \bibinfo{year}{2021 (accessed August, 2021)}\natexlab{}.
\newblock \bibinfo{booktitle}{\emph{{The Machine Learning Reproducibility
  Checklist}}}.
\newblock
\urldef\tempurl%
\url{https://www.cs.mcgill.ca/~jpineau/ReproducibilityChecklist.pdf}
\showURL{%
\tempurl}


\bibitem[Amershi et~al\mbox{.}(2019)]%
        {AmershiSEIP19}
\bibfield{author}{\bibinfo{person}{Saleema Amershi}, \bibinfo{person}{Andrew
  Begel}, \bibinfo{person}{Christian Bird}, \bibinfo{person}{Robert DeLine},
  \bibinfo{person}{Harald~C. Gall}, \bibinfo{person}{Ece Kamar},
  \bibinfo{person}{Nachiappan Nagappan}, \bibinfo{person}{Besmira Nushi}, {and}
  \bibinfo{person}{Thomas Zimmermann}.} \bibinfo{year}{2019}\natexlab{}.
\newblock \showarticletitle{Software engineering for machine learning: a case
  study}. In \bibinfo{booktitle}{\emph{Proceedings of the 41st International
  Conference on Software Engineering: Software Engineering in Practice, {ICSE}
  {(SEIP)} 2019, Montreal, QC, Canada, May 25-31, 2019}},
  \bibfield{editor}{\bibinfo{person}{Helen Sharp} {and} \bibinfo{person}{Mike
  Whalen}} (Eds.). \bibinfo{publisher}{{IEEE} / {ACM}},
  \bibinfo{pages}{291--300}.
\newblock


\bibitem[Arbel et~al\mbox{.}(2023)]%
        {arbel2023rethinking}
\bibfield{author}{\bibinfo{person}{Michael Arbel}, \bibinfo{person}{Romain
  Menegaux}, {and} \bibinfo{person}{Pierre Wolinski}.}
  \bibinfo{year}{2023}\natexlab{}.
\newblock \showarticletitle{Rethinking Gauss-Newton for learning
  over-parameterized models}.
\newblock \bibinfo{journal}{\emph{Advances in neural information processing
  systems}} (\bibinfo{year}{2023}).
\newblock


\bibitem[Barrak et~al\mbox{.}(2021)]%
        {BarrakSANER21}
\bibfield{author}{\bibinfo{person}{Amine Barrak}, \bibinfo{person}{Ellis~E.
  Eghan}, {and} \bibinfo{person}{Bram Adams}.} \bibinfo{year}{2021}\natexlab{}.
\newblock \showarticletitle{On the Co-evolution of {ML} Pipelines and Source
  Code - Empirical Study of {DVC} Projects}. In \bibinfo{booktitle}{\emph{28th
  {IEEE} International Conference on Software Analysis, Evolution and
  Reengineering, {SANER} 2021, Honolulu, HI, USA, March 9-12, 2021}}.
  \bibinfo{publisher}{{IEEE}}, \bibinfo{pages}{422--433}.
\newblock


\bibitem[Biewald(2020)]%
        {biewald_wandb_2020}
\bibfield{author}{\bibinfo{person}{Lukas Biewald}.}
  \bibinfo{year}{2020}\natexlab{}.
\newblock \bibinfo{title}{Experiment Tracking with Weights and Biases}.
\newblock
\newblock
\urldef\tempurl%
\url{https://www.wandb.com/}
\showURL{%
\tempurl}
\newblock
\shownote{Software available from wandb.com}.


\bibitem[Bradbury et~al\mbox{.}(2018)]%
        {jax2018github}
\bibfield{author}{\bibinfo{person}{James Bradbury}, \bibinfo{person}{Roy
  Frostig}, \bibinfo{person}{Peter Hawkins}, \bibinfo{person}{Matthew~James
  Johnson}, \bibinfo{person}{Chris Leary}, \bibinfo{person}{Dougal Maclaurin},
  \bibinfo{person}{George Necula}, \bibinfo{person}{Adam Paszke},
  \bibinfo{person}{Jake Vander{P}las}, \bibinfo{person}{Skye
  Wanderman-{M}ilne}, {and} \bibinfo{person}{Qiao Zhang}.}
  \bibinfo{year}{2018}\natexlab{}.
\newblock \bibinfo{booktitle}{\emph{{JAX}: composable transformations of
  {P}ython+{N}um{P}y programs}}.
\newblock
\urldef\tempurl%
\url{http://github.com/google/jax}
\showURL{%
\tempurl}


\bibitem[Chen et~al\mbox{.}(2022)]%
        {Chen:2022d}
\bibfield{author}{\bibinfo{person}{Boyuan Chen}, \bibinfo{person}{Mingzhi Wen},
  \bibinfo{person}{Yong Shi}, \bibinfo{person}{Dayi Lin},
  \bibinfo{person}{Gopi~Krishnan Rajbahadur}, {and} \bibinfo{person}{Zhen~Ming
  Jiang}.} \bibinfo{year}{2022}\natexlab{}.
\newblock \showarticletitle{Towards training reproducible deep learning
  models}. In \bibinfo{booktitle}{\emph{Proceedings of the 44th International
  Conference on Software Engineering}}. \bibinfo{pages}{2202--2214}.
\newblock


\bibitem[Ferenc et~al\mbox{.}(2020)]%
        {ferenc2020deep}
\bibfield{author}{\bibinfo{person}{Rudolf Ferenc}, \bibinfo{person}{Tam{\'a}s
  Viszkok}, \bibinfo{person}{Tam{\'a}s Aladics}, \bibinfo{person}{Judit
  J{\'a}sz}, {and} \bibinfo{person}{P{\'e}ter Heged{\H{u}}s}.}
  \bibinfo{year}{2020}\natexlab{}.
\newblock \showarticletitle{Deep-water framework: The Swiss army knife of
  humans working with machine learning models}.
\newblock \bibinfo{journal}{\emph{SoftwareX}}  \bibinfo{volume}{12}
  (\bibinfo{year}{2020}), \bibinfo{pages}{100551}.
\newblock


\bibitem[Gebru et~al\mbox{.}(2018)]%
        {GebruDatasheet18}
\bibfield{author}{\bibinfo{person}{Timnit Gebru}, \bibinfo{person}{Jamie
  Morgenstern}, \bibinfo{person}{Briana Vecchione},
  \bibinfo{person}{Jennifer~Wortman Vaughan}, \bibinfo{person}{Hanna~M.
  Wallach}, \bibinfo{person}{Hal~Daum{\'{e}} III}, {and} \bibinfo{person}{Kate
  Crawford}.} \bibinfo{year}{2018}\natexlab{}.
\newblock \showarticletitle{Datasheets for Datasets}.
\newblock \bibinfo{journal}{\emph{CoRR}}  \bibinfo{volume}{abs/1803.09010}
  (\bibinfo{year}{2018}).
\newblock
\showeprint[arxiv]{1803.09010}
\urldef\tempurl%
\url{http://arxiv.org/abs/1803.09010}
\showURL{%
\tempurl}


\bibitem[Gharibi et~al\mbox{.}(2019a)]%
        {gharibi2019automated}
\bibfield{author}{\bibinfo{person}{Gharib Gharibi}, \bibinfo{person}{Vijay
  Walunj}, \bibinfo{person}{Rakan Alanazi}, \bibinfo{person}{Sirisha Rella},
  {and} \bibinfo{person}{Yugyung Lee}.} \bibinfo{year}{2019}\natexlab{a}.
\newblock \showarticletitle{Automated management of deep learning experiments}.
  In \bibinfo{booktitle}{\emph{Proceedings of the 3rd International Workshop on
  Data Management for End-to-End Machine Learning}}. \bibinfo{pages}{1--4}.
\newblock


\bibitem[Gharibi et~al\mbox{.}(2019b)]%
        {gharibi2019modelkb}
\bibfield{author}{\bibinfo{person}{Gharib Gharibi}, \bibinfo{person}{Vijay
  Walunj}, \bibinfo{person}{Sirisha Rella}, {and} \bibinfo{person}{Yugyung
  Lee}.} \bibinfo{year}{2019}\natexlab{b}.
\newblock \showarticletitle{Modelkb: towards automated management of the
  modeling lifecycle in deep learning}. In \bibinfo{booktitle}{\emph{2019
  IEEE/ACM 7th International Workshop on Realizing Artificial Intelligence
  Synergies in Software Engineering (RAISE)}}. IEEE, \bibinfo{pages}{28--34}.
\newblock


\bibitem[Grigorescu et~al\mbox{.}(2020)]%
        {GrigorescuJFR20}
\bibfield{author}{\bibinfo{person}{Sorin~Mihai Grigorescu},
  \bibinfo{person}{Bogdan Trasnea}, \bibinfo{person}{Tiberiu~T. Cocias}, {and}
  \bibinfo{person}{Gigel Macesanu}.} \bibinfo{year}{2020}\natexlab{}.
\newblock \showarticletitle{A survey of deep learning techniques for autonomous
  driving}.
\newblock \bibinfo{journal}{\emph{J. Field Robotics}} \bibinfo{volume}{37},
  \bibinfo{number}{3} (\bibinfo{year}{2020}), \bibinfo{pages}{362--386}.
\newblock


\bibitem[Hill et~al\mbox{.}(2016)]%
        {hill2016trials}
\bibfield{author}{\bibinfo{person}{Charles Hill}, \bibinfo{person}{Rachel
  Bellamy}, \bibinfo{person}{Thomas Erickson}, {and} \bibinfo{person}{Margaret
  Burnett}.} \bibinfo{year}{2016}\natexlab{}.
\newblock \showarticletitle{Trials and tribulations of developers of
  intelligent systems: A field study}. In \bibinfo{booktitle}{\emph{2016 IEEE
  Symposium on Visual Languages and Human-Centric Computing (VL/HCC)}}. IEEE,
  \bibinfo{pages}{162--170}.
\newblock


\bibitem[Hutson(2018)]%
        {HutsonScience18}
\bibfield{author}{\bibinfo{person}{Matthew Hutson}.}
  \bibinfo{year}{2018}\natexlab{}.
\newblock \showarticletitle{Artificial intelligence faces reproducibility
  crisis}.
\newblock \bibinfo{journal}{\emph{Science (New York, N.Y.)}}
  \bibinfo{volume}{359} (\bibinfo{date}{02} \bibinfo{year}{2018}),
  \bibinfo{pages}{725--726}.
\newblock
\urldef\tempurl%
\url{https://doi.org/10.1126/science.359.6377.725}
\showDOI{\tempurl}


\bibitem[Idowu et~al\mbox{.}(2021)]%
        {IdowuSEIP21}
\bibfield{author}{\bibinfo{person}{Samuel Idowu}, \bibinfo{person}{Daniel
  Str{\"{u}}ber}, {and} \bibinfo{person}{Thorsten Berger}.}
  \bibinfo{year}{2021}\natexlab{}.
\newblock \showarticletitle{Asset Management in Machine Learning: {A} Survey}.
  In \bibinfo{booktitle}{\emph{43rd {IEEE/ACM} International Conference on
  Software Engineering: Software Engineering in Practice, {ICSE} {(SEIP)} 2021,
  Madrid, Spain, May 25-28, 2021}}. \bibinfo{publisher}{{IEEE}},
  \bibinfo{pages}{51--60}.
\newblock


\bibitem[Idowu et~al\mbox{.}(2022)]%
        {Idowu:2022}
\bibfield{author}{\bibinfo{person}{Samuel Idowu}, \bibinfo{person}{Daniel
  Str{\"u}ber}, {and} \bibinfo{person}{Thorsten Berger}.}
  \bibinfo{year}{2022}\natexlab{}.
\newblock \showarticletitle{Asset management in machine learning:
  State-of-research and state-of-practice}.
\newblock \bibinfo{journal}{\emph{Comput. Surveys}} \bibinfo{volume}{55},
  \bibinfo{number}{7} (\bibinfo{year}{2022}), \bibinfo{pages}{1--35}.
\newblock


\bibitem[Isdahl and Gundersen(2019)]%
        {isdahl2019out}
\bibfield{author}{\bibinfo{person}{Richard Isdahl} {and}
  \bibinfo{person}{Odd~Erik Gundersen}.} \bibinfo{year}{2019}\natexlab{}.
\newblock \showarticletitle{Out-of-the-box reproducibility: A survey of machine
  learning platforms}. In \bibinfo{booktitle}{\emph{2019 15th international
  conference on eScience (eScience)}}. IEEE, \bibinfo{pages}{86--95}.
\newblock


\bibitem[Mitchell et~al\mbox{.}(2019)]%
        {MitchellFAT19}
\bibfield{author}{\bibinfo{person}{Margaret Mitchell}, \bibinfo{person}{Simone
  Wu}, \bibinfo{person}{Andrew Zaldivar}, \bibinfo{person}{Parker Barnes},
  \bibinfo{person}{Lucy Vasserman}, \bibinfo{person}{Ben Hutchinson},
  \bibinfo{person}{Elena Spitzer}, \bibinfo{person}{Inioluwa~Deborah Raji},
  {and} \bibinfo{person}{Timnit Gebru}.} \bibinfo{year}{2019}\natexlab{}.
\newblock \showarticletitle{Model Cards for Model Reporting}. In
  \bibinfo{booktitle}{\emph{Proceedings of the Conference on Fairness,
  Accountability, and Transparency, FAT* 2019, Atlanta, GA, USA, January 29-31,
  2019}}, \bibfield{editor}{\bibinfo{person}{danah boyd} {and}
  \bibinfo{person}{Jamie~H. Morgenstern}} (Eds.). \bibinfo{publisher}{{ACM}},
  \bibinfo{pages}{220--229}.
\newblock


\bibitem[Mora-Cantallops et~al\mbox{.}(2021)]%
        {mora2021traceability}
\bibfield{author}{\bibinfo{person}{Mar{\c{c}}al Mora-Cantallops},
  \bibinfo{person}{Salvador S{\'a}nchez-Alonso}, \bibinfo{person}{Elena
  Garc{\'\i}a-Barriocanal}, {and} \bibinfo{person}{Miguel-Angel Sicilia}.}
  \bibinfo{year}{2021}\natexlab{}.
\newblock \showarticletitle{Traceability for trustworthy ai: A review of models
  and tools}.
\newblock \bibinfo{journal}{\emph{Big Data and Cognitive Computing}}
  \bibinfo{volume}{5}, \bibinfo{number}{2} (\bibinfo{year}{2021}),
  \bibinfo{pages}{20}.
\newblock


\bibitem[Ormenisan et~al\mbox{.}(2020)]%
        {ormenisan2020implicit}
\bibfield{author}{\bibinfo{person}{Alexandru~A Ormenisan},
  \bibinfo{person}{Mahmoud Ismail}, \bibinfo{person}{Seif Haridi}, {and}
  \bibinfo{person}{Jim Dowling}.} \bibinfo{year}{2020}\natexlab{}.
\newblock \showarticletitle{Implicit provenance for machine learning
  artifacts}.
\newblock \bibinfo{journal}{\emph{Proceedings of MLSys}}  \bibinfo{volume}{20}
  (\bibinfo{year}{2020}).
\newblock


\bibitem[pandas~development team(2020)]%
        {reback2020pandas}
\bibfield{author}{\bibinfo{person}{The pandas~development team}.}
  \bibinfo{year}{2020}\natexlab{}.
\newblock \bibinfo{booktitle}{\emph{pandas-dev/pandas: Pandas}}.
\newblock
\urldef\tempurl%
\url{https://doi.org/10.5281/zenodo.3509134}
\showDOI{\tempurl}


\bibitem[Paszke et~al\mbox{.}(2019)]%
        {paszke_pytorch_2019}
\bibfield{author}{\bibinfo{person}{Adam Paszke}, \bibinfo{person}{Sam Gross},
  \bibinfo{person}{Francisco Massa}, \bibinfo{person}{Adam Lerer},
  \bibinfo{person}{James Bradbury}, \bibinfo{person}{Gregory Chanan},
  \bibinfo{person}{Trevor Killeen}, \bibinfo{person}{Zeming Lin},
  \bibinfo{person}{Natalia Gimelshein}, \bibinfo{person}{Luca Antiga},
  \bibinfo{person}{Alban Desmaison}, \bibinfo{person}{Andreas Köpf},
  \bibinfo{person}{Edward Yang}, \bibinfo{person}{Zach DeVito},
  \bibinfo{person}{Martin Raison}, \bibinfo{person}{Alykhan Tejani},
  \bibinfo{person}{Sasank Chilamkurthy}, \bibinfo{person}{Benoit Steiner},
  \bibinfo{person}{Lu Fang}, \bibinfo{person}{Junjie Bai}, {and}
  \bibinfo{person}{Soumith Chintala}.} \bibinfo{year}{2019}\natexlab{}.
\newblock \bibinfo{title}{{PyTorch}: {An} {Imperative} {Style},
  {High}-{Performance} {Deep} {Learning} {Library}}.
\newblock
\newblock
\newblock
\shownote{\_eprint: 1912.01703}.


\bibitem[Pedregosa et~al\mbox{.}(2011)]%
        {pedregosa2011scikit}
\bibfield{author}{\bibinfo{person}{Fabian Pedregosa}, \bibinfo{person}{Ga{\"e}l
  Varoquaux}, \bibinfo{person}{Alexandre Gramfort}, \bibinfo{person}{Vincent
  Michel}, \bibinfo{person}{Bertrand Thirion}, \bibinfo{person}{Olivier
  Grisel}, \bibinfo{person}{Mathieu Blondel}, \bibinfo{person}{Peter
  Prettenhofer}, \bibinfo{person}{Ron Weiss}, \bibinfo{person}{Vincent
  Dubourg}, {et~al\mbox{.}}} \bibinfo{year}{2011}\natexlab{}.
\newblock \showarticletitle{Scikit-learn: Machine learning in Python}.
\newblock \bibinfo{journal}{\emph{Journal of machine learning research}}
  \bibinfo{volume}{12}, \bibinfo{number}{Oct} (\bibinfo{year}{2011}),
  \bibinfo{pages}{2825--2830}.
\newblock


\bibitem[Rasti et~al\mbox{.}(2023)]%
        {rasti_image_2023}
\bibfield{author}{\bibinfo{person}{Behnood Rasti}, \bibinfo{person}{Alexandre
  Zouaoui}, \bibinfo{person}{Julien Mairal}, {and} \bibinfo{person}{Jocelyn
  Chanussot}.} \bibinfo{year}{2023}\natexlab{}.
\newblock \showarticletitle{Image {Processing} and {Machine} {Learning} for
  {Hyperspectral} {Unmixing}: {An} {Overview} and the {HySUPP} {Python}
  {Package}}.
\newblock \bibinfo{journal}{\emph{arXiv preprint arXiv:2308.09375}}
  (\bibinfo{year}{2023}).
\newblock


\bibitem[Serban et~al\mbox{.}(2020)]%
        {Serban:2020}
\bibfield{author}{\bibinfo{person}{Alex Serban}, \bibinfo{person}{Koen van~der
  Blom}, \bibinfo{person}{Holger Hoos}, {and} \bibinfo{person}{Joost Visser}.}
  \bibinfo{year}{2020}\natexlab{}.
\newblock \showarticletitle{Adoption and effects of software engineering best
  practices in machine learning}. In \bibinfo{booktitle}{\emph{Proceedings of
  the 14th ACM/IEEE International Symposium on Empirical Software Engineering
  and Measurement (ESEM)}}. \bibinfo{pages}{1--12}.
\newblock


\bibitem[Sinha et~al\mbox{.}(2023)]%
        {Sinha:2023}
\bibfield{author}{\bibinfo{person}{Koustuv Sinha}, \bibinfo{person}{Maurits
  Bleeker}, \bibinfo{person}{Samarth Bhargav}, \bibinfo{person}{Jessica~Zosa
  Forde}, \bibinfo{person}{Sharath~Chandra Raparthy}, \bibinfo{person}{Jesse
  Dodge}, \bibinfo{person}{Joelle Pineau}, {and} \bibinfo{person}{Robert
  Stojnic}.} \bibinfo{year}{2023}\natexlab{}.
\newblock \showarticletitle{{ML Reproducibility Challenge 2022}}.
\newblock \bibinfo{journal}{\emph{ReScience C}} \bibinfo{volume}{9},
  \bibinfo{number}{2} (\bibinfo{date}{July} \bibinfo{year}{2023}).
\newblock


\bibitem[Sodhani et~al\mbox{.}(2021)]%
        {sodhani_multi-task_2021}
\bibfield{author}{\bibinfo{person}{Shagun Sodhani}, \bibinfo{person}{Amy
  Zhang}, {and} \bibinfo{person}{Joelle Pineau}.}
  \bibinfo{year}{2021}\natexlab{}.
\newblock \showarticletitle{Multi-task reinforcement learning with
  context-based representations}. In \bibinfo{booktitle}{\emph{International
  {Conference} on {Machine} {Learning}}}. \bibinfo{publisher}{PMLR},
  \bibinfo{pages}{9767--9779}.
\newblock


\bibitem[Suvorov et~al\mbox{.}(2022)]%
        {suvorov_resolution-robust_2022}
\bibfield{author}{\bibinfo{person}{Roman Suvorov}, \bibinfo{person}{Elizaveta
  Logacheva}, \bibinfo{person}{Anton Mashikhin}, \bibinfo{person}{Anastasia
  Remizova}, \bibinfo{person}{Arsenii Ashukha}, \bibinfo{person}{Aleksei
  Silvestrov}, \bibinfo{person}{Naejin Kong}, \bibinfo{person}{Harshith Goka},
  \bibinfo{person}{Kiwoong Park}, {and} \bibinfo{person}{Victor Lempitsky}.}
  \bibinfo{year}{2022}\natexlab{}.
\newblock \showarticletitle{Resolution-robust large mask inpainting with
  fourier convolutions}. In \bibinfo{booktitle}{\emph{Proceedings of the
  {IEEE}/{CVF} winter conference on applications of computer vision}}.
  \bibinfo{pages}{2149--2159}.
\newblock


\bibitem[Takamoto et~al\mbox{.}(2022)]%
        {takamoto_pdebench_2022}
\bibfield{author}{\bibinfo{person}{Makoto Takamoto}, \bibinfo{person}{Timothy
  Praditia}, \bibinfo{person}{Raphael Leiteritz}, \bibinfo{person}{Daniel
  MacKinlay}, \bibinfo{person}{Francesco Alesiani}, \bibinfo{person}{Dirk
  Pflüger}, {and} \bibinfo{person}{Mathias Niepert}.}
  \bibinfo{year}{2022}\natexlab{}.
\newblock \showarticletitle{{PDEBench}: {An} extensive benchmark for scientific
  machine learning}.
\newblock \bibinfo{journal}{\emph{Advances in Neural Information Processing
  Systems}}  \bibinfo{volume}{35} (\bibinfo{year}{2022}),
  \bibinfo{pages}{1596--1611}.
\newblock


\bibitem[Tatman et~al\mbox{.}(2018)]%
        {tatman2018practical}
\bibfield{author}{\bibinfo{person}{Rachael Tatman}, \bibinfo{person}{Jake
  VanderPlas}, {and} \bibinfo{person}{Sohier Dane}.}
  \bibinfo{year}{2018}\natexlab{}.
\newblock \bibinfo{title}{A practical taxonomy of reproducibility for machine
  learning research}.
\newblock
\newblock


\bibitem[Team et~al\mbox{.}(2023)]%
        {team2023gemini}
\bibfield{author}{\bibinfo{person}{Gemini Team}, \bibinfo{person}{Rohan Anil},
  \bibinfo{person}{Sebastian Borgeaud}, \bibinfo{person}{Yonghui Wu},
  \bibinfo{person}{Jean-Baptiste Alayrac}, \bibinfo{person}{Jiahui Yu},
  \bibinfo{person}{Radu Soricut}, \bibinfo{person}{Johan Schalkwyk},
  \bibinfo{person}{Andrew~M Dai}, \bibinfo{person}{Anja Hauth},
  {et~al\mbox{.}}} \bibinfo{year}{2023}\natexlab{}.
\newblock \showarticletitle{Gemini: a family of highly capable multimodal
  models}.
\newblock \bibinfo{journal}{\emph{arXiv preprint arXiv:2312.11805}}
  (\bibinfo{year}{2023}).
\newblock


\bibitem[Tsay et~al\mbox{.}(2018)]%
        {tsay2018runway}
\bibfield{author}{\bibinfo{person}{Jason Tsay}, \bibinfo{person}{Todd Mummert},
  \bibinfo{person}{Norman Bobroff}, \bibinfo{person}{Alan Braz},
  \bibinfo{person}{Peter Westerink}, {and} \bibinfo{person}{Martin Hirzel}.}
  \bibinfo{year}{2018}\natexlab{}.
\newblock \showarticletitle{Runway: machine learning model experiment
  management tool}. In \bibinfo{booktitle}{\emph{Conference on systems and
  machine learning (sysML)}}.
\newblock


\bibitem[Vartak et~al\mbox{.}(2016)]%
        {vartak2016modeldb}
\bibfield{author}{\bibinfo{person}{Manasi Vartak}, \bibinfo{person}{Harihar
  Subramanyam}, \bibinfo{person}{Wei-En Lee}, \bibinfo{person}{Srinidhi
  Viswanathan}, \bibinfo{person}{Saadiyah Husnoo}, \bibinfo{person}{Samuel
  Madden}, {and} \bibinfo{person}{Matei Zaharia}.}
  \bibinfo{year}{2016}\natexlab{}.
\newblock \showarticletitle{ModelDB: a system for machine learning model
  management}. In \bibinfo{booktitle}{\emph{Proceedings of the Workshop on
  Human-In-the-Loop Data Analytics}}. \bibinfo{pages}{1--3}.
\newblock


\bibitem[Wei{\ss}gerber and Granitzer(2019)]%
        {weissgerber2019mapping}
\bibfield{author}{\bibinfo{person}{Thomas Wei{\ss}gerber} {and}
  \bibinfo{person}{Michael Granitzer}.} \bibinfo{year}{2019}\natexlab{}.
\newblock \showarticletitle{Mapping platforms into a new open science model for
  machine learning}.
\newblock \bibinfo{journal}{\emph{it-Information Technology}}
  \bibinfo{volume}{61}, \bibinfo{number}{4} (\bibinfo{year}{2019}),
  \bibinfo{pages}{197--208}.
\newblock


\bibitem[Yadan(2019)]%
        {yadan_hydra_2019}
\bibfield{author}{\bibinfo{person}{Omry Yadan}.}
  \bibinfo{year}{2019}\natexlab{}.
\newblock \bibinfo{title}{Hydra - {A} framework for elegantly configuring
  complex applications}.
\newblock
\newblock
\urldef\tempurl%
\url{https://github.com/facebookresearch/hydra}
\showURL{%
\tempurl}


\bibitem[Zaharia et~al\mbox{.}(2018)]%
        {chen_developments_2020}
\bibfield{author}{\bibinfo{person}{Matei Zaharia}, \bibinfo{person}{Andrew
  Chen}, \bibinfo{person}{Aaron Davidson}, \bibinfo{person}{Ali Ghodsi},
  \bibinfo{person}{Sue~Ann Hong}, \bibinfo{person}{Andy Konwinski},
  \bibinfo{person}{Siddharth Murching}, \bibinfo{person}{Tomas Nykodym},
  \bibinfo{person}{Paul Ogilvie}, \bibinfo{person}{Mani Parkhe},
  {et~al\mbox{.}}} \bibinfo{year}{2018}\natexlab{}.
\newblock \showarticletitle{Accelerating the machine learning lifecycle with
  MLflow.}
\newblock \bibinfo{journal}{\emph{IEEE Data Eng. Bull.}} \bibinfo{volume}{41},
  \bibinfo{number}{4} (\bibinfo{year}{2018}), \bibinfo{pages}{39--45}.
\newblock


\end{thebibliography}

\end{document}